\documentclass{article}

 \usepackage[final]{nips_2018}

\usepackage{titlesec}

\usepackage[utf8]{inputenc} 
\usepackage[T1]{fontenc}    
\usepackage{hyperref}       
\usepackage{url}            
\usepackage{booktabs}       
\usepackage{amsfonts}       
\usepackage{nicefrac}       
\usepackage{microtype}      
\usepackage{graphicx}
\usepackage{algorithm,algorithmic}
\usepackage{amssymb,mathrsfs,epsfig,epstopdf,amsthm}
\usepackage{mathtools}
\usepackage{physics}
\usepackage{wrapfig}
\title{Mixture of Regression Experts in fMRI Encoding}

\author{Subba Reddy Oota*\textsuperscript{1}, 
Adithya Avvaru\textsuperscript{1},
Naresh Manwani\textsuperscript{1}, 
Raju S. Bapi\textsuperscript{1,2} \\
\textsuperscript{1} IIIT Hyderabad, 
\textsuperscript{2} University of Hyderabad \\
\{oota.subba,adithya.avvaru\}@students.iiit.ac.in, \{naresh.manwani,raju.bapi\}@iiit.ac.in\\}

\begin{document}
\def \xx {\mathbf{x}}
\def \yy {\mathbf{y}}
\maketitle

\begin{abstract}
fMRI semantic category understanding using linguistic encoding models attempt to learn a forward mapping that relates stimuli to the corresponding brain activation. 
Classical encoding models use linear multi-variate methods to predict the brain activation (all voxels) given the stimulus. 
However, these methods essentially assume multiple regions as one large uniform region or several independent regions, ignoring connections among them. 
In this paper, we present a mixture of experts-based model where a group of experts captures brain activity patterns related to  particular regions of interest (ROI) and also show the discrimination across different experts.
The model is trained word stimuli encoded as 25-dimensional feature vectors as input and the corresponding brain responses as output. Given a new word (25-dimensional feature vector), it predicts the entire brain activation as the linear combination of multiple experts' brain activations. 
We argue that each expert learns a certain region of brain activations corresponding to its category of words, which solves the problem of identifying the regions with a simple encoding model. 
We showcase that proposed mixture of experts-based model indeed learns region-based experts to predict the brain activations with high spatial accuracy.
\end{abstract}

\section{Introduction}
In recent years, the use of both linear and non-linear multivariate encoding/decoding approaches for analyzing fMRI brain activity have become increasingly popular~\citep{mitchell2008predicting,naselaris2011encoding,mesgarani2014phonetic,di2015low,pereira2018toward}.
An encoding model that predicts brain activity in response to stimuli is important for neuroscientists who can use the model predictions to investigate and test hypotheses about the transformation from stimulus to brain response in patients. Recent approaches for modeling fMRI data use word embedding representations to build encoding systems~\citep{oota2018fmri,abnar2018experiential}.
However, all these models do not have a principled way of learning regions specializing in particular category of stimuli. Instead, they simply predict the voxel intensity values either for the whole brain or for a pre-selected set of voxels. 
In particular, interpreting the non-linear multivariate models can be difficult due to capturing of unsuspected and enigmatic representations~\citep{benjamin2017modern,kording2018roles}. 

Mixture of experts (MoE) models, proposed by Jordan's group~\citep{jordan1995convergence} offer an interesting choice for the problem of learning distinct models for different regions of the input space. MoE were applied to MRI data to investigate the complex patterns of brain changes associated with non-pathological and pathological processes, such as the effects of growth, aging, injury or a  disease~\citep{kim2010bayesian,eavani2016capturing}. MoEs have great potential for use in medical diagnostics to diagnose a variety of clinical conditions such as depression, Alzheimer's dementia etc. ~\citep{yao2009hierarchical} proposed a Hidden Conditional Random Framework (HCRF) inspired from a mixture of experts model to make predictions in all ROIs which are interconnected.  

In this paper, we present a mixture of regression experts (MoRE) model that predicts the brain activity associated with concrete nouns as stimuli. The theory underlying this model is that each expert learns to cover a certain brain region of interest (set of voxels that are significantly activated) based on the category of words that are represented by the model. Finally, fMRI signals are modeled as a linear combination of all expert predictions corresponding to the gate probabilities of each expert. We present an experimental evidence showing that the best encoding model is achieved with the mixture of regression experts rather than using a simple linear/non-linear model alone.

\section{fMRI Encoding: Mixture of Regression Experts (MoRE) Approach}
\label{Approach}

\begin{figure*}[t]
  \centering
  \includegraphics[width=0.8\linewidth]{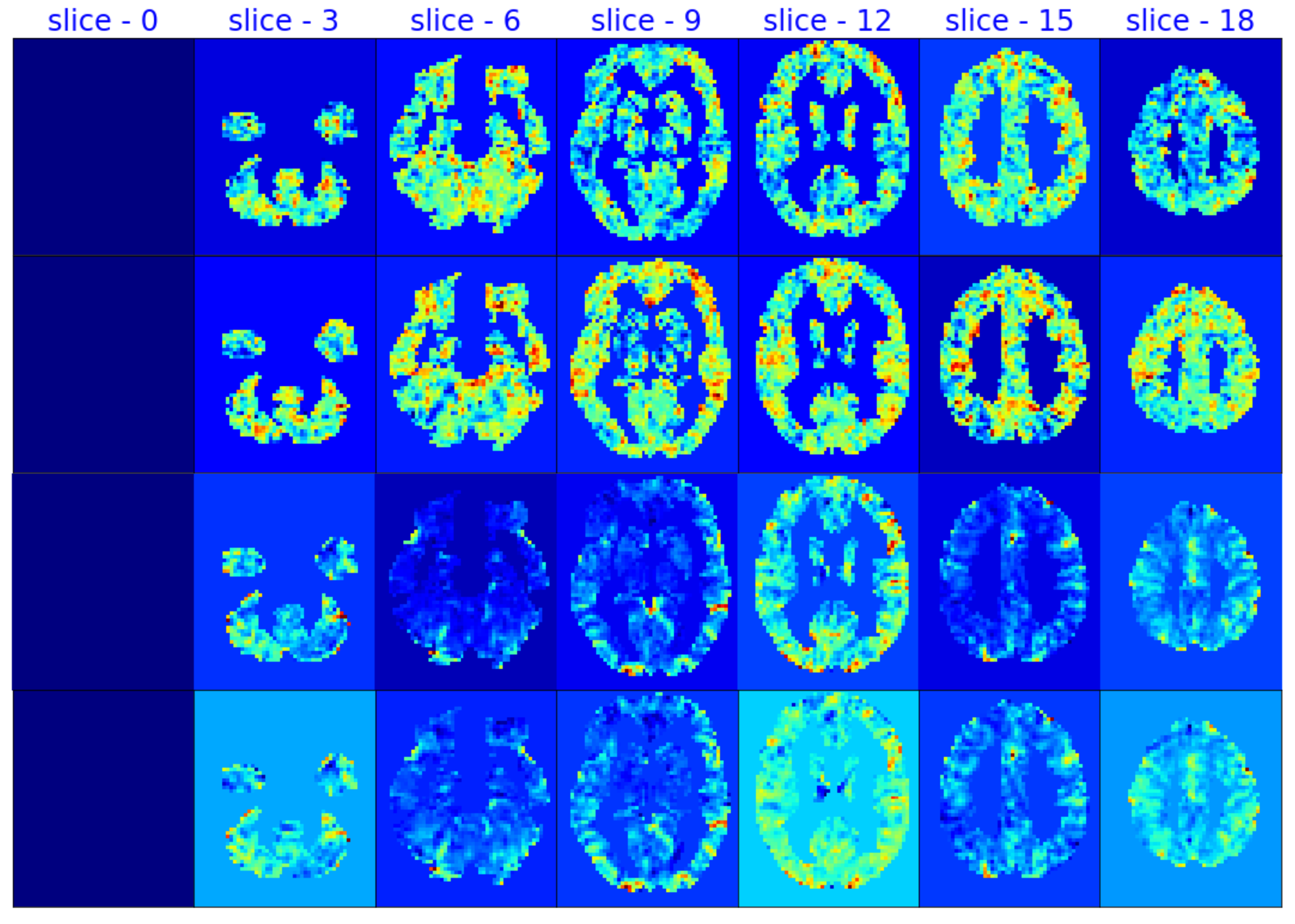}
  \caption{The sequence of slices, for the word ``Apartment'' after converting voxel activation per subject into 23 slices, (i) in actual brain activation  (top row), (ii) for the model trained with mixture of experts model (second row from top), (iii) for the model trained with multilayer perceptron model (third row from top), (iv) for the model trained with ridge regression (bottom row).}
  \label{fig:brainresponse}
\end{figure*}

Current encoding methods attempt to learn either weights in case of linear models~\citep{mitchell2008predicting,pereira2018toward} or complex representations in non-linear models~\citep{oota2018fmri}. In the next sections, we discuss the proposed MoRE approach and our enhancements.
\vspace{-0.3cm}
\subsection{Architecture: }
We use a mixture of experts-based encoder model, whose architecture is inspired from~\citet{jordan1995convergence}. The mixture of experts architecture is composed of a gating network and several expert networks, each of which solves a function approximation problem over a local region of the input space. 

Let $D=\{X,Y\}$ denote the data where $X=\{x^{n}\}_{n=1}^{N}$ is the input, $Y=\{y^{n}\}_{n=1}^{N}$ is the target, $K$ is number of experts and $N$ is the number of training points. Also, let $\Theta=\{\theta_{0},W_{e},\Sigma_{e}\}$ denote the set of parameters where $\theta_{0}$ is the set of gate parameters, $W_{e}$ is the set of expert parameters, and $\Sigma_{e}$ is the set of expert co-variance matrices.

The mixtures of experts model is based on the following conditional mixture
\begin{equation}
p(\yy|\xx) = \sum_{j=1}^{K}P(j|\xx)p(\yy|\xx,\theta_{j}) = \sum_{j=1}^{K}g_{j}(\xx,\theta_{0})p(\yy|\xx,\theta_{j})
\end{equation}
where $g_j(\xx,\theta_0)$ is the gate's rating, i.e., the probability of the $j^{th}$ expert, given $\xx$ and $p(\yy|\xx,\theta_{j})$ denotes the density function for the output vector associated with the $j^{th}$ expert, i.e., the probability of $j^{th}$ expert generating $\yy$ given $\xx$.

Here, we choose $p(\yy|\xx,\theta_{j})$ as multivariate Gaussian probability density for each of the experts, denoted by.
\begin{equation}
p(\yy|\xx,W_j,\Sigma_j) = \frac{1}{(2\pi)^{m/2}|\Sigma_j|^{1/2}}\exp\left(-\frac{1}{2}(\yy-W_j\xx)^{T}\Sigma_j^{-1}(\yy-W_j\xx)\right)
\end{equation}
where $W_j \in \mathbb{R}^{m \times n}$ is the weight matrix and $\Sigma_{j} \in \mathbb{R}^{m\times m}$ is the covariance matrix associated with the $j^{th}$ expert. In our problem, we consider two different cases of $\Sigma_j$. 
\begin{enumerate}
\item $\Sigma_j = \sigma_j^{2}I,\;\forall j \in [K]$  and
\item $\Sigma_j = diag(\sigma_{j,1}^2,\sigma_{j,2}^2,\ldots,\sigma_{j,m}^2),\;\forall j \in [K]$.
\end{enumerate}
Thus, we assume that the components of the output vector ${\bf y} \in \mathbb{R}^m$ are statistically independent of each other. We use this assumption to make the model very simple and reduce the number of overall parameters. This assumption will also make the algorithm computationally less expensive.

Assuming $\Sigma_j = diag(\sigma_{j,1}^2,\sigma_{j,2}^2,\ldots,\sigma_{j,m}^2)$, we rewrite the conditional probability density model for $j^{th}$ expert as follows:
\begin{equation}
P({\bf y}|{\bf x},W_j,\Sigma_{j}) = \frac{1}{(2\pi)^{m/2}(\sigma_{j,1}\sigma_{j,2}..\sigma_{j,m})}\exp\left(-\sum_{i=1}^m \frac{(y_i-{\bf w}_{j,i}^T\xx)^2}{2\sigma_{j,i}^2} \right)
\end{equation}
where ${\bf w}_{j,i}$ is the $i^{th}$ row of $W_j$.

We use softmax function for $g(\xx,\theta_0)$.
\begin{equation}
g_{j}(\xx,\theta_{0}) = \frac{\exp\left({\bf v}_j^T\xx\right)}{\sum_{i=1}^{K}\exp\left({\bf v}_i^T\xx\right)}
\end{equation}
where ${\bf v}_j \in \mathbb{R}^n,\;\forall j \in [K]$. Thus, $\theta_0=\{{\bf v}_1,\ldots,{\bf v}_K\}$.

\section{Experiments}
\label{headings}

\paragraph{Dataset}
\label{sec:datasets}

\begin{wrapfigure}{r}{5cm} 
    \includegraphics[width=0.35\textwidth]{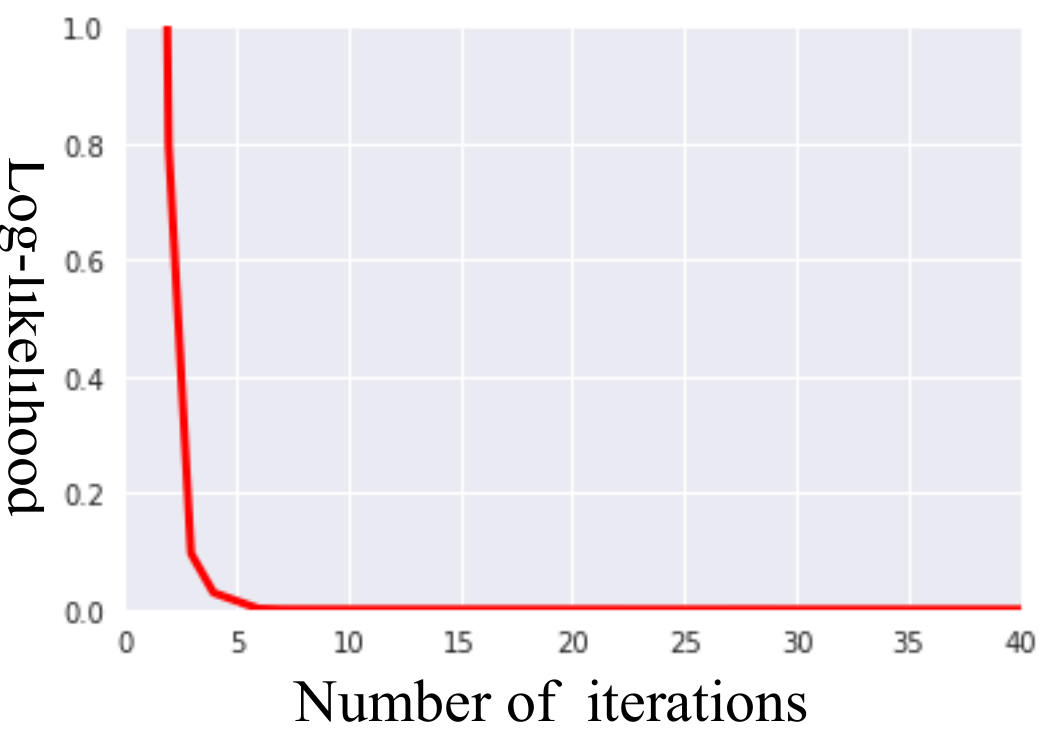}
    \caption{MoRE log-likelihood convergence}
    \label{fig:loglikelihood}
\end{wrapfigure}

We used data from fMRI experiment 1~\citep{mitchell2008predicting}, where authors conducted experiments with multiple subjects by showing different categories of words as stimuli. Each category might correspond to activation of distinct brain regions. 
In the experiment, the target word was presented with a picture that depicted some aspect(s) of the relevant meaning. 
This fMRI dataset was collected from a total of 9 participants. 
For each participant in the experiment, a total set of 60 words (12 categories) were used as stimuli in multi-modal form (word, picture).
The fMRI dataset constitutes $51 \times 61$  voxel windows arranged as 23 slices, per subject per stimulus.
We use the publicly available Mitchell's 25-feature vector data~\citep{mitchell2008predicting} of 60 words as input and the corresponding brain response of each participant (containing 21000 voxels) as output to train the model.

\begin{figure}
  \centering
  \includegraphics[width=0.75\linewidth]{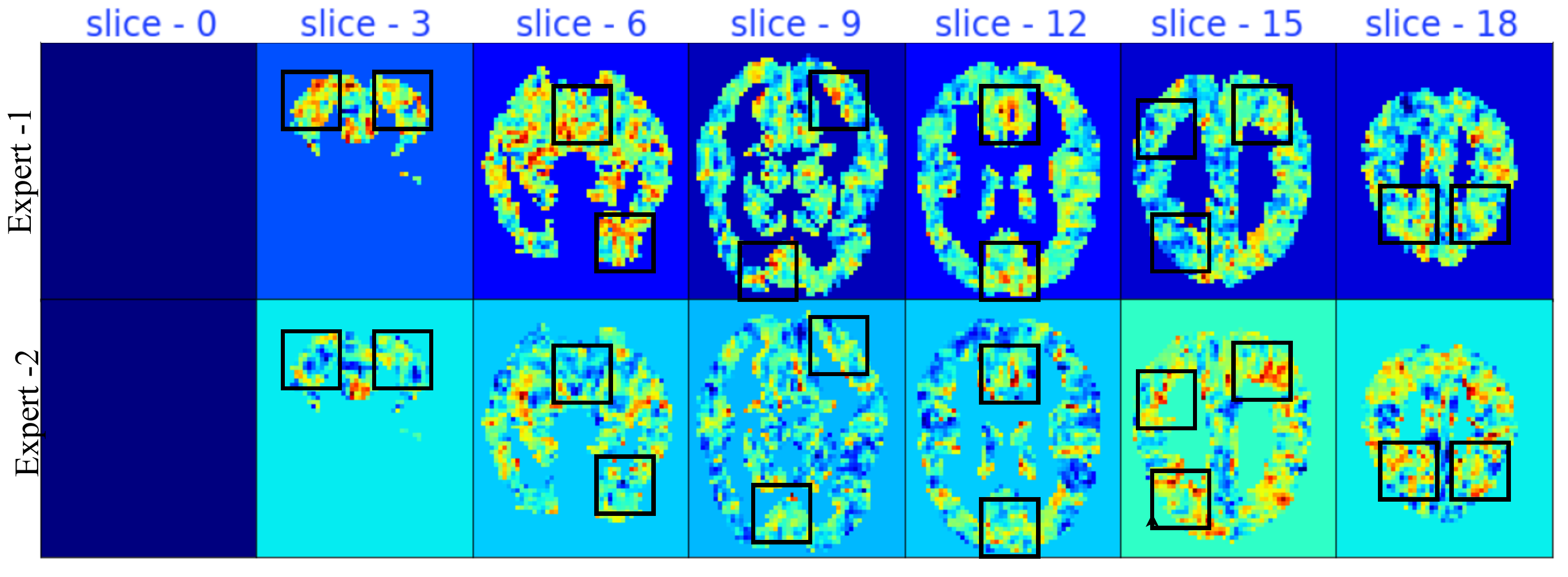}
  \caption{Comparison of brain activity (ROIs) in two experts across slices. The squared box indicates the ROIs' activation differences: i) Expert-1 has higher brain activity in ``cingulate region" and ``Superior and Middle frontal gyrus regions''(top row), ii) ROIs captured in Expert2 are ``occipital lobe" from visual region'', ``Inferior temporal gyrus'' and ``Middle temporal gyrus'' (bottom row)}
  \label{fig:experts}
  \vspace{-0.1cm}
\end{figure}

\paragraph{Selection of Number of Experts: }
\label{sec:experts}
In this paper, we selected the number of experts based on the number of categories considered in the dataset. The dataset is divided into 12 categories in which each category has 5 concrete nouns. The idea is that each expert should master a particular category and the corresponding activation in relevant brain regions. The detailed selecting number expert models using Bayesian Information Criteria provided in Appendix~\ref{BIC}. 

\paragraph{Description of the Experiment}
\label{sec:experiment_description}
Here, we considered Mitchell's feature vector, $\xx$ as input to all the experts and calculated $p(\yy|\xx,\theta_{j})$ where $\yy$ is the corresponding brain responses for that $\xx$. Gating network trains to  pick the expert which gives maximum $g_j(\xx,\theta_0)$. The detailed training methodology is provided in Appendix~\ref{sec:Training_MOE_Appendix}.

\paragraph{Results and Discussion}
\label{sec:results}
Using the approach discussed in Section~\ref{Approach} and Appendix~\ref{sec:Training_MOE_Appendix}, we trained separate mixture of regression experts model for each subject.
The encoding performance was evaluated by training and testing models using different subsets of the 60 concrete words in a 5-fold cross-validation scheme. 
The encoder models were trained until the model reached convergence as shown in Figure~\ref{fig:loglikelihood}.
Figure~\ref{fig:experts} shows the region of activity in two different experts for the word ``Church''.
Across all the slices, there is a considerable number of ROI differences among all the experts. In particular, we depict regions of activation for two experts in Figure~\ref{fig:experts}.  
Figure~\ref{fig:experts} shows that Expert 1 has higher brain activity in ``cingulate region" and ``Superior and Middle frontal gyrus regions''. 
In Expert 2, the ROIs captured are ``Occipital lobe" from visual region'', ``Inferior temporal gyrus'' and ``Middle temporal gyrus''. It appears that while Expert 1 mediates activation in the frontal regions usually involved in decision making, Expert 2 seems to activate regions responsible for visual processing and object recogntion. In addition, we observed some brain activity being common to ROIs across all the experts (results not shown here).
In order to assess the similarity between the actual and predicted activation in each brain slice, we compared slice-wise voxel coordinates and their intensity values.
We measured the model scores using predictive $r^{2}$ score, for voxel $v$ as follows: $r_{v}^{2} = 1 - \frac{||\yy^{v}_{test}-\yy^v_{pred}||^{2}}{||\yy^{v}_{test}-mean(\yy^v_{test})||^{2}}$, where $\yy^{v}_{test}$ is the testing set brain activations, $\yy^v_{pred}$ is the predicted brain activations and mean($\yy^v_{test}$) is the mean activation on the test set.
Table~\ref{results1} depicts subject-wise (9 subjects) $r^{2}$ score comparisons among the mixture of experts model, the model trained with the multilayer perceptron and with the ridge regression model.From Table~\ref{results1}, we observe that $r^{2}$ score for mixture of experts model is better than that of the other two models and the score range indicates that MoRE model typically yields better predictions.

\setlength{\tabcolsep}{3.2pt}
\begin{table}[t]
\centering
\begin{tabular}{|c|c c c c c c c c c|} \hline
&\multicolumn{9}{ c |}{\textbf{Subjects ($r^{2}$ score) }} \\
Method& (1)& (2)& (3) & (4) & (5) &(6) &(7) &(8) & (9)\\ \hline \hline
\textbf{Mixture of Regression Experts}&0.811& 0.813 & 0.806 &0.811&0.814 & 0.815 & 0.813 & 0.808 & 0.813\\ 
Multilayer Perceptron&  0.755 & 0.745 & 0.745 & 0.772 & 0.75 & 0.765 & 0.756 & 0.761&0.751\\ 
Ridge Regression&  0.772& 0.769 &0.758&0.776&0.778&0.782&0.776&0.770&0.771\\ \hline
\end{tabular}
\vspace{0.1cm}
\caption{Subject-wise $r^{2}$ scores for different models}
\label{results1}
\end{table}

\section{Conclusion}
In this paper, we present a mixture of experts-based model where group of experts capture brain activity patterns related to  particular regions of interest (ROIs) and also show the discrimination across different experts.
Different from previous work, the underlying model depicts that each expert trains on certain brain regions of interest (set of voxels that are significantly activated) based on the category of words that are represented by the model.
The key distinction of our work is the utilization of experts model to capture the ROIs based on various categories of input.
In future, we plan to experiment on all paradigms and conduct experiments on other fMRI datasets and identify the number of experts using Bayesian Information Criterion (BIC), with a primary focus on hierarchical mixture of experts at slice-level instead of voxel-level predictions. 

\bibliographystyle{images/nips_2017}
\bibliography{nips_2018}

\appendix
\section{Selecting Number of Experts using BIC}
\label{BIC}
In order to find the number of experts, Bayesian Information Criteria (BIC) is one of the successful measure to approximate the Bayes factor~\citet{kass1995bayes} i.e to find a model that has maximum posterior probability or maximum marginal likelihood. The BIC can be formulated as follows.

\begin{align}
    BIC = d\log(n) - 2\log(p(X|\Theta))
\end{align}

in~\citet{schwarz1978estimating}, where d is the dimensionality of the parameter space, n is the number of data points, $p(X|\Theta)$ denotes the parametric model density. 

\section{Training Mixture of Experts using EM Algorithm}
\label{sec:Training_MOE_Appendix}
Let $N$ be the number of samples, $K$ be the number of experts and  $S=\{(\xx_1,\yy_1),\ldots,(\xx_N,\yy_N)\}$ be the training set where $(\xx^n,\yy^n)\in \mathbb{R}^k\times \mathbb{R}^m,\;\forall n \in [N]$.

The EM algorithm is an iterative method for finding the maximum likelihood (ML) of a probability model in which some random variables are observed and others are hidden. In training the Mixture of Experts, the indicator variables $Z = z_{j}^{n},\; j\in[K],\;n\in [N]$ are introduced to solve the model with the EM algorithm.
\begin{gather*}
z_{j}^{n} =
\begin{cases}
              1, \text{if $\yy^{n}$ is generated from the $j^{th}$ expert}\\    
              \text{0, otherwise}    
\end{cases}
\end{gather*}

With the indicator variables, the likelihood can be written as
\begin{align}
L(\Theta;S;Z) &= P(S, Z|\Theta) = \prod_{n=1}^{N}P(\yy^{n}|\xx^{n}) \nonumber \\
										   &= \prod_{n=1}^{N}\sum_{j=1}^{K}[g_{j}(\xx^{n},\theta_{0})P(\yy^{n}|\xx^{n},W_{j},\Sigma_{j})]^{z_{j}^{n}}
\end{align}

The parameter $\Theta$ can be estimated by the iterations through the E and M steps.

\subsection{E-Step : Calculation of Expectation of log-likelihood}

The EM algorithm is employed to average out $z_{j}$ and maximize the expected complete data log-likelihood $E_{Z}(log P(D, Z|\Theta))$ and $\sum_{j=1}^{K}z_{j}^{n}$ =1, for each $n$. The expectation of the log-likelihood in (5) results in

\begin{align}
Q(\Theta|\Theta^{(p)})  &= E_{Z}(\log P(D, Z|\Theta)) \nonumber\\
                       &= \sum_{n=1}^{N}\sum_{j=1}^{K}h_{j}^{(p)}(n) \big\{log([g_{j}(\xx^{n},\theta_{0})P(\yy^{n}|\xx^{n},W_{j},\Sigma_{j})]) \big\}\nonumber \\
                 &= \sum_{n=1}^{N}\sum_{j=1}^{K}h_{j}^{(p)}(n) \log(g_{j}(\xx^{n},\theta_{0})) + \sum_{n=1}^{N}\sum_{j=1}^{K}h_{j}^{(p)}(n) \log(P(\yy^{n}|\xx^{n},W_{j},\Sigma_{j})))
\end{align}

where $p$ is the iteration index and $h_{j}^{(p)}(n)$ is expectation of indicator variables which is given by
\begin{align}
h_{j}^{(p)}(n) 	&= E[z_{j}^{n}|D] \nonumber \\
				&= P(z_{j}^{n}=1|\xx^{(n)},\yy^{(n)}) \nonumber \\
                & = \frac{g_{j}(\xx^{(n)},\theta_{0})P(\yy^{(n)}|\xx^{(n)},W_{j},\Sigma_{j})}{\sum_{i=1}^{K}g_{i}(\xx^{(n)},\theta_{0})P(\yy^{(n)}|\xx^{(n)},W_{i},\Sigma_{i})} 
\end{align}

\subsection{M-Step : Estimation of new parameters $\Theta^{(p+1)}$}
\begin{itemize}
\item From E-step, we calculated the expectation of log-likelihood $Q$, we now investigate the implementation of M-step by estimating the new parameters. 
\item The M step chooses a parameter $\Theta$ that increases the $Q$ function, the expected value of the complete-data log-likelihood i.e.

\begin{align}
    \Theta^{(p+1)}=\underset{\Theta}{\text{argmax}} \hspace{2pt} Q(\Theta|\Theta^{(p)})
\end{align}

\item An iteration of EM also increases the original log likelihood $L$. That is $L(\Theta^{(p+1)};Y)>L(\Theta^{(p)};Y)$
i.e. the likelihood $L$ increases monotonically along the sequence of parameter estimates generated by an EM algorithm.
\end{itemize}
\paragraph{Update Gate Parameters}
\begin{itemize}
\item From the Equations (4) and (6) 

\begin{equation}
E_{gate} = -\sum_{t=1}^{N}\sum_{j=1}^{K}h_{j}^{(p)}(n)log(g_{j}(\xx^{(n)},\theta_{0})
\end{equation}

\item An interpretation of the cost function in Equation (6) is the cross-entropy of distributing a pattern $\xx$ amongst the expert networks. This cost is minimal if experts are mutually exclusive and increases when experts share a pattern.

\begin{align}
\pdv{Q}{\theta_{0}} &= \sum_{n=1}^{N}\sum_{j=1}^{K}h_{j}^{(p)}(n) \pdv{g_{j}(\xx^{(n)},\theta_{0})}{\theta_{0}}/g_{j}(\xx^{(n)},\theta_{0}) \nonumber\\
 &= \sum_{n=1}^{N}\sum_{j=1}^{K}h_{j}^{(p)}(n) [\pdv{({\bf v}_{j}^{T}\xx^{n})}{\theta_{0}}-\sum_{i=1}^{K}g_{i}(\xx^{(n)},\theta_{0})\pdv{({\bf v}_{i}^{T}\xx^{n})}{\theta_{0}}] \nonumber\\
 &= \sum_{n=1}^{N}\sum_{j=1}^{K}[h_{j}^{(p)}(n)-g_{j}(\xx^{(n)},\theta_{0})]\pdv{({\bf v}_{j}^{T}\xx^{n})}{\theta_{0}}
\end{align}

As the gating network is a specific form of a generalized linear model, the general update for $\theta_{0}$ is written as follows
\begin{equation}
\theta_{0}^{(p+1)} = \theta_{0}^{(p)}+\eta_{g}\pdv{Q}{\theta_{0}}
\end{equation}

\end{itemize}

\begin{itemize}
\item Calculate the log likelihood using both $\theta_{0}^{(p)},\theta_{0}^{(p+1)}$ in Equation (5), and the difference in log likelihood values is calculated as 
\begin{align}
    \triangledown L = L(\Theta^p;S;Z) - L(\Theta^{p+1};S;Z)    
\end{align}

\item If $\triangledown L$ is below a threshold, then new parameters are discarded and MOE model will use old parameters.
\item There is no known closed-form way to find the minimum of cost function, and thus we'll resort to an iterative optimization algorithm such as gradient descent or L-BFGS.
\end{itemize}

\paragraph{Update Expert Parameters}
\begin{itemize}
\item From the Equations (3) and (6) 

\begin{equation}
E_{expert} = -\sum_{n=1}^{N}\sum_{j=1}^{K}h_{j}^{(p)}(n) log(P(\yy^{n}|\xx^{n},W_{j},\Sigma_{j}))
\end{equation}

\item The update equation for expert parameters $W_{j}$ as follows
\end{itemize}

\begin{align}
\pdv{Q}{W_{j}} &= \sum_{n=1}^{N}\sum_{j=1}^{K}h_{j}^{(p)}(n) \pdv{P(\yy^{(n)}|\xx^{(n)},W_{j})}{W_{j}}/P(\yy^{(n)}|\xx^{(n)},W_{j}) \nonumber\\
&= \sum_{n=1}^{N}\sum_{j=1}^{K}h_{j}^{(p)}(n) \pdv{(W_{j}^{T}\xx^{(n)})}{W_{j}}\frac{[\yy^{(n)}-W_{j}^{T}\xx^{(n)}]}{\Sigma_{j}} \nonumber\\
&= \sum_{n=1}^{N}\sum_{j=1}^{K}h_{j}^{(p)}(n) \xx^{(n)}\frac{[\yy^{(n)}-W_{j}^{T}\xx^{(n)}]}{\Sigma_{j}}
\end{align}

\begin{equation}
W_{j}^{(p+1)} = W_{j}^{(p)}+\eta_{j}\pdv{Q}{W_{j}}
\end{equation}

\begin{itemize}
\item Calculate the log likelihood using both $W_{j}^{(p)}, W_{j}^{(p+1)}$ in Equation (5), and the difference in log likelihood values is calculated as 
\begin{align}
    \triangledown L = L(\Theta^p;S;Z) - L(\Theta^{p+1};S;Z)    
\end{align}

\item If $\triangledown L$ is below a threshold, then new parameters are discarded and MOE model will use old parameters.
\end{itemize}

\paragraph{Update Expert Covariance Matrix}
\begin{itemize}
\item The update equation for covariance matrix as follows.
\begin{equation}
\Sigma_{j}^{(p+1)} = \frac{1}{\sum_{n=1}^{N}h_{j}^{(p)}(n)}\sum_{n=1}^{N}h_{j}^{(p)}(n)[\yy^{(n)}-W_{j}^{T}\xx^{(n)}]^{2}
\end{equation}

\item Calculate the log likelihood using both $\Sigma_{j}^{(p)},\Sigma_{j}^{(p+1)}$ in Equation (5), and the difference in log likelihood values is calculated as 
\begin{align}
    \triangledown L = L(\Theta^p;S;Z) - L(\Theta^{p+1};S;Z)    
\end{align}

\item If $\triangledown L$ is below a threshold, then new parameters are discarded and MOE model will use old parameters.

\end{itemize}

\end{document}